\documentclass{article}
\usepackage{spconf,amsmath,graphicx}

\usepackage{color, soul, multirow}
\usepackage{amssymb}
\usepackage{subfig}
\usepackage{bm}
\usepackage{verbatim}
\usepackage{cite}
\usepackage{amssymb}
\usepackage{pifont}
\usepackage{tabularx}
\usepackage{lipsum}
\usepackage{float}
\usepackage{booktabs}

\DeclareMathOperator*{\argmax}{arg\,max}

\title{Improved Neural Language Model Fusion \\
for Streaming Recurrent Neural Network Transducer}
\name{
Suyoun Kim, Yuan Shangguan, Jay Mahadeokar, Antoine Bruguier
}
\address{\vspace{-0.9cm}\\
\emph{Christian Fuegen, Michael L. Seltzer, Duc Le} \\ 
\vspace{0.1cm}\\
Facebook AI, USA}

\begin{document}
%
\maketitle
\begin{abstract}

Recurrent Neural Network Transducer (RNN-T), like most end-to-end speech recognition model architectures, has an implicit neural network language model (NNLM) and cannot easily leverage unpaired text data during training. Previous work has proposed various fusion methods to incorporate external NNLMs into end-to-end ASR to address this weakness. In this paper, we propose extensions to these techniques that allow RNN-T to exploit external NNLMs during both training and inference time,  resulting in 13-18\% relative Word Error Rate improvement on Librispeech compared to strong baselines. Furthermore, our methods do not incur extra algorithmic latency and allow for flexible plug-and-play of different NNLMs without re-training. We also share in-depth analysis to better understand the benefits of the different NNLM fusion methods. Our work provides a reliable technique for leveraging unpaired text data to significantly improve RNN-T while keeping the system streamable, flexible, and lightweight.

\end{abstract}
\begin{keywords}
RNN-T, language model fusion, streaming end-to-end speech recognition, leveraging unpaired text
\end{keywords}
\section{Introduction}
\label{sec:intro}


Recurrent Neural Network Transducer (RNN-T) \cite{Graves12transduction,Prabhavalkar17,battenberg2017exploring,he2019streaming, li2019improving} has become one of the most popular model architectures for on-device streaming automatic speech recognition (ASR) over the last few years. Compared to traditional hybrid ASR systems, RNN-T is much more compact due to the lack of external n-gram language models (LMs) and decision trees. Compared to other end-to-end ASR approaches such as encoder-decoder with attention \cite{chorowski2015attention,chan2016listen,bahdanau2016end,kim2017joint,Chiu18}, RNN-T is easier to stream and generally works better in low-latency scenarios where the entire utterance is not available up front and partial decoding results need to be emitted during decoding.

Although RNN-T has the advantage of streamability over encoder-decoder based ASR models, a recent study \cite{weinstein2020stateless} found that the prediction network of RNN-T, often thought of as an implicit LM, shows poor results in modeling long-range linguistic information. In addition, the model's end-to-end nature makes it difficult to leverage unpaired text data to further improve performance. Many recent works have investigated methods for using text-only data to improve encoder-decoder based ASR models, including fusion with an external neural network LM (NNLM) \cite{gulcehre2015using, kannan2018analysis, sriram2017cold, shan2019component, toshniwal2018comparison}. There has been comparatively limited work on NNLM fusion for RNN-T with streaming constraints, where shallow fusion remains the most popular and effective technique \cite{kannan2018analysis,McDermott19}. A recent study \cite{weinstein2020stateless} tried pre-training the prediction network of RNN-T with unpaired text, but did not get any WER improvement.

In this work, we explore NNLM fusion methods for RNN-T that are applied on-the-fly during first pass decoding, thus avoiding additional algorithmic latency and keeping the model latency low. We propose extensions to the original cold NNLM fusion to increase its flexibility and effectiveness within the RNN-T framework. Our combined cold and shallow NNLM fusion method achieves \textbf{13-18\%} relative WER improvement on the widely used Librispeech dataset over our strong baselines. In addition, our method allows for flexible plug-and-play of different NNLMs without the need for re-training, which could be very useful for rapid domain adaptation and dynamically adjusting to resource constraints. Lastly, we provide in-depth analysis to better understand the benefits of the different NNLM fusion methods.

\section{NNLM fusion for RNN-T}
\label{sec:method}
\subsection{RNN-T Overview}

RNN-T \cite{Graves12transduction} consists of three major sub-networks: encoder, predictor, and joiner. The encoder transforms an input sequence of audio feature vectors $\mathbf{x} = (x_1, \dots, x_T)$ into a sequence of acoustic embeddings $\mathbf{h}^{enc}$:

\begin{equation}
    \mathbf{h}^{enc} = f^{enc}(\mathbf{x}) = (h^{enc}_1, \dots, h^{enc}_{T'})   
\end{equation}

\noindent where $T'$ may be different from $T$. The predictor, which is analogous to an LM, transforms a sequence of previous tokens $(y_1, \dots, y_{u-1})$ into an embedding vector $h^{pred}_u$:

\begin{equation}
    h^{pred}_u = f^{pred}(y_1, \dots, y_{u-1})
\label{eq:predictor}
\end{equation}

\noindent Finally, the joiner combines the encoder embedding $h^{enc}_t$ and predictor embedding $h^{pred}_u$ to estimate the logits $z_{t,u}$:

\begin{equation}
    z_{t,u} = f^{join}(h^{enc}_t, h^{pred}_u)
\label{eq:joiner}
\end{equation}
\begin{equation}
    P(.|x_1, \dots, x_t, y_1, \dots, y_{u-1}) = \text{softmax}(z_{t,u})
\end{equation}

Additional details on the RNN-T training objective and decoding procedure can be found in \cite{Graves12transduction}.

\subsection{Shallow Fusion}
\label{sec:shallow}

Shallow fusion is the most popular technique for combining RNN-T with an external NNLM trained on text-only data \cite{kannan2018analysis}. In shallow fusion, the NNLM is incorporated via log-linear interpolation at inference time, and the decoding problem of finding the best hypothesis $y^*$ becomes:

\begin{equation}
    y^* = \argmax_y \log P_\text{RNNT}(y|x) + \lambda \log P_{LM} (y)
\end{equation}

\noindent where $\lambda$ is a hyperparameter that controls the relative importance of the external NNLM ($\lambda = 0$ corresponds to normal RNN-T decoding without shallow fusion). Note that unlike encoder-decoder models, we cannot directly interpolate the log probability of RNN-T’s joiner output and NNLM output at each output time step because RNN-T allows emission of blank symbols which are not modeled in external NNLMs. In our implementation, we interpolate external NNLM scores with RNN-T scores during beam search when the model outputs a non-blank output symbol. This interpolation happens on-the-fly and the overall system remains streamable.


\subsection{Cold Fusion}
\label{sec:cold}

One limitation of shallow fusion is that the external NNLM is only applied during inference. Cold fusion \cite{gulcehre2015using,sriram2017cold} is a method originally proposed for encoder-decoder models where a pre-trained external NNLM is fused directly into the decoder network by combining their hidden states during training time. Similar to the decoder network of encoder-decoder models, the prediction network of RNN-T is analogous to an LM. Our proposed cold fusion method for RNN-T extends Equation (\ref{eq:predictor}) to combine the predictor embedding and NNLM output as follows: 
\begin{align}
    s_u^{LM} &= \text{softmax}(z^{LM}_u) \\
    \label{eq:bottle}
    h_u^{LM} &= f^{LM}(s_u^{LM})\\
    \label{eq:gate1}
    g_u &= \text{sigmoid}(W_g [h_u^{pred}; h_u^{LM}] + b_g) \\
    \label{eq:gate2}
    h_u^{CF} &= f^{CF}(g_u \odot [h_u^{pred} ; h_u^{LM}])
\end{align}

\noindent where $z^{LM}_u$ is the external NNLM's predicted logits over non-blank output symbols for the next time step given a sequence of previously emitted tokens $(y_1, \dots, y_{u-1})$, $f^{LM}$ is the LM projection network, and $f^{CF}$ is the combined projection network. The final embedding $h_u^{CF}$ has the same dimension as $h_u^{pred}$ and replaces the latter in Equation (\ref{eq:joiner}).

Some of the key differences between our cold fusion approach and the original cold fusion formulation are:
\begin{enumerate}
    \item We adopt an iterative training procedure instead of training the network from scratch. This means the RNN-T and NNLM are first pre-trained separately; the cold fusion RNN-T is then finetuned with the frozen NNLM for a few epochs. We will show in Section \ref{ssec:iterative} that this technique is crucial for cold fusion to work.
    \item We use the NNLM's logits $z_u^{LM}$ instead of its hidden state. This allows us to swap in a different NNLM during inference without re-training the whole network. Example scenarios where this ability could be useful are switching to a lightweight NNLM where computation power is limited, or plugging in a domain-specific NNLM for improved accuracy. We will showcase this flexibility in Section \ref{ssec:lm_swap}.
    \item We apply a fine gating mechanism on top of the concatenated predictor and NNLM output, as shown in Equation (\ref{eq:gate1}) and (\ref{eq:gate2}). This gating mechanism can increase the predictor network's modeling power by using multiplicative interaction\cite{kim2018towards}, together with additional linguistic information from the external NNLM.
\end{enumerate}

With cold fusion, the system remains streamable since NNLM scores are combined on-the-fly in first pass decoding.



\begin{table*}[t]
\centering
        \begin{tabular}{rrrccc}
        \toprule
        \textbf{Encoder Arch.} & \textbf{Params} & \textbf{Lookahead} & \textbf{LM Fusion} & \textbf{test-clean} & \textbf{test-other} \\
        \midrule
        \midrule
        BLSTM \cite{Chiu2020} & $\sim$126M & $\inf$ (non-streaming) & None & 3.2 & 7.8 \\
        \midrule
        \multirow{2}{*}{Transformer \cite{zhang2020transformer}} & \multirow{2}{*}{139M} & 30ms (feature stacking) & \multirow{2}{*}{None} & 4.2 & 11.3\\
         & & $\sim$1.1s (2 frames/layer) & & 3.0 & 7.7\\
        \midrule
        \midrule
        \multirow{4}{*}{LSTM (ours)} & \multirow{4}{*}{65M} & \multirow{4}{*}{100ms (feature stacking)} & None & 4.0 & 10.1 \\
         & & & \texttt{SF} & 3.3 & 8.6  \\
         & & & \texttt{CF} & 3.8 & 9.4 \\
         & & & \texttt{SF+CF} & \textbf{3.3} & \textbf{8.3} \\
        \midrule
        \multirow{4}{*}{LC-BLSTM (ours)} & \multirow{4}{*}{99M} & \multirow{4}{*}{240ms (right context)} & None & 3.2 & 8.0\\
         & & & \texttt{SF} & 2.8 & 7.0\\
         & & & \texttt{CF} & 3.0 & 7.6\\
         & & & \texttt{SF+CF} & \textbf{2.8} & \textbf{6.8}\\
        \bottomrule
        \end{tabular}
\caption{Librispeech WER comparison between vanilla RNN-T baselines, shallow fusion (\texttt{SF}), cold fusion (\texttt{CF}), combined fusion (\texttt{SF+CF}), and relevant published results using sequence transducers.}
\label{tab:res}
\end{table*}

\section{Experiments}

\textbf{Data:} We conduct experiments on the widely used Librispeech \cite{panayotov2015librispeech} dataset which consists of 960 hours of labeled speech and an additional text-only corpus containing 810M words. We use 80-dim globally z-normalized logMel filterbank coefficients as acoustic features, derived from 25ms FFT windows with a 10ms frame shift. We apply the Librispeech Double policy without time warping from SpecAugment \cite{park2019specaugment} during training. We also perform speed perturbation \cite{ko2015audio} of the training data and produce three versions of each audio with speed factors 0.9, 1.0, and 1.1; as a result, the training data size is tripled. The output targets are 5000 unigram WordPieces \cite{Kudo2018SubWord} generated by the SentencePiece toolkit \cite{kudo2018sentencepiece}, plus an additional blank symbol.

\textbf{RNN-T Encoder:} We consider two of the most popular streaming encoder architectures for RNN-T, Long-Short Term Memory (LSTM) and Latency Controlled Bidirectional LSTM (LC-BLSTM) \cite{zhang2016highway} as our baselines in this work. The LSTM encoder stacks 11 contiguous feature frames as input, consists of eight layers with 1024 cells each, and has a 640-dim linear projection after every LSTM layer. The LC-BLSTM encoder works on single feature frames without stacking, has 24-frame lookahead (i.e., 240ms), 120-frame chunk size, and comprises eight layers with 640 cells each. Both encoders subsample the input by a factor of four and produce 1024-dim embeddings. In both networks, the predictor contains two LSTM layers with 512 cells each, and the joiner has a single linear layer. The total trainable parameters are 65M (LSTM) and 99M (LC-BLSTM). We train the models for 80 epochs using Adam \cite{kingma2014adam}; the learning rate is fixed at 0.0004 for the first 60 epochs, then drops by a factor of 0.8 after every subsequent epoch.


\textbf{External NNLM:} We employ a 4-layer LSTM network with 2048 cells in each layer, interleaved with 640-dim linear projection, totaling 53M trainable parameters. We train the model on the 810M text-only corpus (broken down into WordPieces) with Cross Entropy loss for 40 epochs using the Adam optimizer \cite{kingma2014adam}. The learning rate is fixed at 0.0004 for the first 25 epochs, then drops by a factor of 0.8 after every subsequent epoch. This NNLM will be the basis for our shallow fusion and cold fusion experiments.

\textbf{Cold Fusion:} The LM projection network $f^{LM}$ contains a 256-dim bottleneck layer, followed by linear projection into 1024 dimensions. The combined projection network $f^{CF}$ consists of a single 1024-dim linear projection layer. These cold fusion-specific components add 7.8M trainable parameters in total. We freeze the NNLM parameters, bootstrap the RNN-T from baseline models, and finetune the network for 10 epochs with a 0.0005 learning rate for the first 3 epochs, then decays by a factor of 0.6 after every epoch. For fair comparison, we also tried finetuning the baseline models for 10 more epochs, but did not obtain better results.

\textbf{Decoding:} We use a beam size of 15 for all experiments. The shallow fusion interpolation weight $\lambda$ ranges between 0.2 and 0.5 based on tuning results on development sets. The optimal $\lambda$ is smaller when cold fusion is combined with shallow fusion, likely because the NNLM scores are already implicit within the cold fusion RNN-T scores. Furthermore, we can reuse the NNLM logits for both fusion methods, thus incurring minimal computational overhead when combining them.

\section{Results and Discussion}
\subsection{WER Overview}

Table \ref{tab:res} shows the WER results of our proposed approaches, together with some relevant baselines from published works that also utilized sequence transducers. The main motivation for including these published results is to demonstrate that our baseline model's numbers are very competitive. We are not aiming to outperform the Librispeech state-of-the-art in this paper, which typically entails using non-streamable encoder architectures (e.g., full-context transformer) as well as second pass rescoring on full utterances, whereas the focus of our work is on streaming first-pass decoding.

Both shallow fusion (\texttt{SF}) and cold fusion (\texttt{CF}) significantly improve over the vanilla baselines, with \texttt{SF} giving better results than \texttt{CF}. It is unclear why \texttt{CF} underperforms compared to \texttt{SF}, even though the NNLM is incorporated directly in training. We hypothesize that \texttt{SF} is able to distribute the probability mass more evenly to different WordPieces via direct interpolation, whereas the spiky nature of RNN-T scores limits the impact of \texttt{CF}. We will verify this hypothesis in future work. Combining \texttt{SF} and \texttt{CF} results in further improvement on the more challenging \texttt{test-other} split, producing an overall WER reduction of \textbf{13-18\%} over the baselines.

\subsection{Importance of Iterative Training}
\label{ssec:iterative}

    \begin{table}[tb]
    \centering
            \begin{tabular}{rcc}
            \toprule
            \textbf{Model} & \textbf{test-clean} & \textbf{test-other} \\
            \midrule
            \midrule
            LSTM \texttt{CF} (from scratch) & \multicolumn{2}{c}{Did Not Converge} \\
            LSTM \texttt{CF} (iterative) & 3.8 & 9.4 \\
            \midrule
            LC-BLSTM \texttt{CF} (from scratch) & 3.2 & 8.4 \\
            LC-BLSTM \texttt{CF} (iterative) & 3.0 & 7.6 \\
            \bottomrule
            \end{tabular}
    \caption{Effect of iterative training on cold fusion (\texttt{CF}).}
    \label{tab:iterative}
    \end{table}

Table \ref{tab:iterative} shows that training cold fusion RNN-T from scratch fails to yield WER improvement; we also observed that the model has difficulties converging, especially with an LSTM encoder. It is therefore crucial to adopt the iterative training approach, i.e., bootstrapping from a well-trained RNN-T model. We hypothesize that the strong signal from the external NNLM makes the model rely less on RNN-T, thus the latter becomes under-trained. Conversely, most parts of the network are already well-trained in the iterative scenario (only a few linear layers for cold fusion need to be trained from scratch), and the finetuning process mainly teaches the model how to integrate the available signals.

\subsection{Flexible NNLM Swapping}
\label{ssec:lm_swap}

The use of NNLM's logits $z^{LM}_u$ allows for flexible plug-and-play of different NNLMs during inference without re-training the whole network. This flexibility can be useful in many scenarios. For example, we only have to train the model once and simply plug in different NNLMs for different devices or surfaces depending on their resource constraints. Table \ref{tab:swap} shows an example where we swap out the original NNLM (53M parameters) with a lightweight version (15M parameters). The swapped-in LM works out of the box and gives similar results as using it directly in cold fusion training.

In cases where we have strong apriori knowledge about the input audio, such as domain information, we could plug in a domain-specific NNLM to obtain better recognition results. Table \ref{tab:swap} illustrates this ability where we swap out the original 53M NNLM with an oracle 15M NNLM trained on the combined \texttt{test-clean} and \texttt{test-other} splits, resulting in massive WER reduction.

    \begin{table}[tb]
    \centering
            \begin{tabular}{rcc}
            \toprule
            \textbf{Model} & \textbf{test-clean} & \textbf{test-other} \\
            \midrule
            \midrule
            LSTM \texttt{CF} (15M LM) & 4.0 & 9.7 \\
            LSTM \texttt{CF} (53M LM) & 3.8 & 9.4 \\
            + swap 15M LM & 3.9 & 9.6 \\
            + swap 15M oracle LM & 2.2 & 5.8 \\
            \midrule
            LC-BLSTM \texttt{CF} (15M LM) & 3.1 & 7.8 \\
            LC-BLSTM \texttt{CF} (53M LM) & 3.0 & 7.6 \\
            + swap 15M LM & 3.1 & 7.8 \\
            + swap 15M oracle LM & 1.9 & 4.9 \\
            \bottomrule
            \end{tabular}
    \caption{Swapping NNLM in cold fusion without re-training.}
    \label{tab:swap}
    \end{table}

\subsection{When Does NNLM Fusion Help?}
\label{sec:analysis}

\begin{table}[tb]
\centering
    \begin{tabular}{rrrrr}
    \toprule
     & \texttt{SF} & \texttt{CF} & \texttt{SF+CF} \\
    \midrule
    \midrule
    \multicolumn{1}{l}{\textbf{\emph{Average Length}}} & & & \\
    Short (8 words)   & 6.7\%  & 3.5\% & 9.5\%  \\
    Medium (16 words) & 14.3\% & 6.7\% & 17.2\%  \\
    Long (34 words)   & 18.7\% & 7.2\% & 20.7\% \\
    \midrule
    \multicolumn{1}{l}{\textbf{\emph{Word Type (\# Utts)}}} & & & \\
    Common (3.8K)      & 17.4\% & 7.9\% & 20.2\% \\
    Fixed by LM (1.5K) & 14.3\% & 5.4\% & 16.5\% \\
    Rare/OOV (332)     &  9.1\% & 3.8\% & 11.0\% \\
    \bottomrule
    \end{tabular}
\caption{Breakdown of relative WER reduction of shallow fusion (\texttt{SF}), cold fusion (\texttt{CF}), and combined fusion (\texttt{SF+CF}) compared to vanilla baselines. Analysis is done on LSTM's \texttt{test-clean} and \texttt{test-other} results.}
\label{tab:werr_breakdown}
\end{table}

We first analyze WER improvement as a function of utterance length. We split utterances in the evaluation set into three chunks: Short, Medium, and Long. Each chunk represents a third of the utterances in the evaluation set, containing around 1.85K utterances. As shown in Table \ref{tab:werr_breakdown} (first section), the improvement provided by all fusion methods increases as the utterance becomes longer. This implies that NNLMs can better model long-range linguistic information and compensate for the known weakness of RNN-T's prediction network \cite{weinstein2020stateless}.

Next we analyze the relation between the rare/out-of-vocabulary (OOV) word issue and NNLM fusion improvement. We define rare/OOV word as a word that appears less than 10 times in the acoustic training transcription. We split utterances in the evaluation set into three chunks: (1) \texttt{Common} - utterances that have no rare/OOV word, (2) \texttt{Fixed by LM} - utterances that had rare/OOV word, but the issue was fixed if we include LM unpaired text data, and (3) \texttt{Rare/OOV} - utterances that still have at least one rare/OOV word after including LM unpaired text data. As shown in Table \ref{tab:werr_breakdown} (second section), the improvement provided by NNLM fusion increases when the unpaired text data are able to mitigate the rare/OOV word issue. This suggests that the increased coverage of the unpaired text corpus plays a crucial role in NNLM fusion's effectiveness.



\vspace{0.4cm}        
\section{Conclusion and Future Work}

In this work, we proposed external NNLM fusion methods for RNN-T models capable of leveraging unpaired text data in both training and decoding.  Our methods are applied on-the-fly during first pass decoding, thus do not adversely impact algorithmic latency and the models remain streamable. Moreover, we showed that iterative training is crucial for getting cold fusion to work and we can obtain complementary benefits from combining both shallow and cold fusion.

For future work, we plan to investigate ways to close the gap between cold fusion and shallow fusion, compare our first pass fusion methods with second pass LM rescoring, and further leverage the flexibility of NNLM swapping in cold fusion and apply it to on-the-fly domain adaptation.

\small
\bibliographystyle{IEEEbib}
\bibliography{refs}

\end{document}